\documentclass[letterpaper]{article}
\usepackage{aaai2026}
\usepackage{natbib}
\usepackage{times}
\usepackage{helvet}
\usepackage{tcolorbox}
\tcbuselibrary{listings}
\usepackage{xcolor,soul}

\usepackage{courier}
\usepackage{graphicx}
\usepackage{amsmath,amssymb}

\usepackage{afterpage}
\usepackage{adjustbox}
\usepackage{graphicx}
\usepackage{url}
\usepackage{algpseudocode}
\usepackage{algorithm}
\usepackage{booktabs}
\usepackage{booktabs}
\usepackage{multirow}
\usepackage{listings}
\usepackage{xcolor}
\usepackage{amsmath}
\usepackage{tikz,siunitx}

\definecolor{myblue}{HTML}{407cb9}
\definecolor{myred}{HTML}{e26062}
\definecolor{mygreen}{HTML}{61a658}

\title{MeLA: A Metacognitive LLM-Driven Architecture for Automatic Heuristic Design}

\author{
    Zishang Qiu\textsuperscript{\rm 1}, 
    Xinan Chen\textsuperscript{\rm 1}\thanks{Corresponding author.},
    Long Chen\textsuperscript{\rm 2}, 
    and Ruibin Bai \textsuperscript{\rm 1}\\
    \textsuperscript{\rm 1}School of Computer Science, University of Nottingham Ningbo China, Ningbo, China \\
    \textsuperscript{\rm 2}College of Teacher Education, Zhejiang Normal University, Jinhua, China\\
    \texttt{xinan.chen@nottingham.edu.com}
}

\usepackage{listings}  % 必须添加
\usepackage{xcolor}    % 用于代码高亮
\usepackage{bibentry}

\begin{document}
\nocopyright
\maketitle

\begin{abstract}

This paper introduces MeLA, a Metacognitive LLM-Driven Architecture that presents a new paradigm for Automatic Heuristic Design (AHD). Traditional evolutionary methods operate directly on heuristic code; in contrast, MeLA evolves the instructional prompts used to guide a Large Language Model (LLM) in generating these heuristics. This process of "prompt evolution" is driven by a novel metacognitive framework where the system analyzes performance feedback to systematically refine its generative strategy. MeLA's architecture integrates a problem analyzer to construct an initial strategic prompt, an error diagnosis system to repair faulty code, and a metacognitive search engine that iteratively optimizes the prompt based on heuristic effectiveness. In comprehensive experiments across both benchmark and real-world problems, MeLA consistently generates more effective and robust heuristics, significantly outperforming state-of-the-art methods. Ultimately, this research demonstrates the profound potential of using cognitive science as a blueprint for AI architecture, revealing that by enabling an LLM to metacognitively regulate its problem-solving process, we unlock a more robust and interpretable path to AHD.

\end{abstract}

\begin{links}
    \link{Code/Dataset/Result}{https://github.com/Qzs1335/MeLA}
\end{links}

\section{Introduction}
The quest for high-performance heuristics is a central and enduring challenge in computational intelligence. These specialized algorithms are the engines driving progress in complex optimization, from logistics to drug discovery. Historically, powerful metaheuristics like SCSO \cite{seyyedabbasi2023sand}, WO \cite{han2024walrus}, and SOA \cite{givi2023skill} were the products of a manual, artisanal process requiring deep human expertise. While these algorithms can perform exceptionally well on the specific problems they were designed for, their performance often degrades when applied to different types of problems or even to variations of the original problem. This limitation is a direct consequence of the No Free Lunch (NFL) theorem \cite{wolpert1997no}, which posits that no single heuristic can be universally optimal across all possible problems. Consequently, the reliance on human-driven design for specific problem instances has become a critical bottleneck, hindering the scale and speed of scientific and industrial innovation.

To break this impasse, the field of Automatic Heuristic Design (AHD) emerged, aiming to automate the creation of heuristics. While foundational AHD paradigms like Hyper-Heuristics (HHs) \cite{pillay2018hyper} and Genetic Programming (GP) \cite{burke2007automatic} were a conceptual leap forward, their creative potential is fundamentally capped. By design, they are tethered to human guidance, operating within search spaces defined by expert-supplied components or primitives \cite{mei2022explainable}. This inherent limitation prevents them from achieving true design autonomy, leaving the grand challenge of fully automated discovery unsolved.

The advent of Large Language Models (LLMs) introduced a paradigm shift. Recognizing their potential, the latest research frontier fuses LLMs with the directed pressure of evolutionary computation \cite{zhang2024understanding}. Pioneering systems like FunSearch \cite{romera2024mathematical}, EoH \cite{liu2024evolution}, and ReEvo \cite{ye2024reevo} have demonstrated remarkable success by evolving heuristics, discovering solutions that rival human-engineered ones. However, this paper argues that their focus on evolving \textit{heuristics} is a source of profound limitations. The paradigm's core flaw is its reliance on population-level dynamics, which treats the LLM as a static generator rather than an adaptive learner. By failing to internalize the principles of what constitutes a superior heuristic, this process inherently limits both the peak performance and the generalizability of the discovered strategies.

\begin{figure}[htbp]
    \centering
    \includegraphics[width=8.4cm]{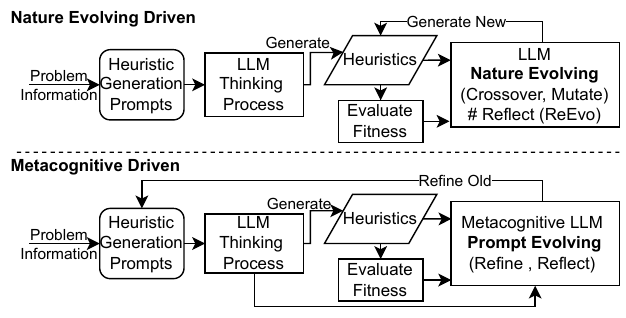} % 缩小到80%
    \caption{Nature Evolving Driven \& Metacognitive Driven Heuristic Generation Architecture}
    \label{compare}
\end{figure}

This work posits that the true point of leverage is not the final heuristic, but the generative \textit{reasoning process} that creates it. To this end, we introduce \textbf{Prompt Evolution}, a new AHD paradigm that fundamentally differs from the natural selection model used by prior work. As illustrated in \textbf{Figure 1}, instead of evolving a population of heuristic codes, our architecture evolves the problem-solving strategy itself, which is encoded in the LLM's guiding prompt. This evolution is driven by a \textbf{metacognitive search engine} inspired by cognitive science \cite{martinez2006metacognition}. This engine assesses the entire causal chain—from the strategic "thinking process" embodied in a prompt to the empirical performance of the heuristic it produces—and uses this analysis to systematically refine the prompt for the next generation of thinking.

This focus on the reasoning process inherently improves the quality and generality of the heuristics generated. However, we recognize that achieving robust performance in practice requires addressing challenges that extend beyond high-level strategy. This is particularly true for complex, ill-defined real-world problems, such as Adaptive Curriculum Sequencing (ACS) \cite{prates2019learning} or Wireless Sensor Network (WSN) deployment \cite{chen2024optimizing}. The ambiguity and situational dependencies of these problems frequently cause prior architectures to fail by generating syntactically or logically flawed code. To overcome these practical hurdles, MeLA is engineered with two novel support mechanisms: an \textbf{Automated Problem Analyzer} that formulates a rich problem description directly from source code, obviating the need for manual input, and an \textbf{Error Diagnosis System} that autonomously detects and rectifies programming flaws in the generated heuristics. These components grant MeLA a degree of autonomy and reliability essential for real-world application.

In summary, this paper introduces MeLA, a metacognitive LLM-driven architecture that advances the state of the art in AHD. The primary contributions are:
\begin{enumerate}
    \item \textbf{A New AHD Paradigm:} We propose and validate \textit{Prompt Evolution}, which focuses on refining the LLM's reasoning process rather than heuristics, leading to more stable and generalizable performance.
    \item \textbf{A Metacognitive Search Engine:} We introduce a novel search mechanism, inspired by metacognition, that enables the architecture to self-reflect and strategically improve its prompt-based problem-solving approach.
    \item \textbf{Robustness for Real-World Application:} We equip the architecture with automated problem analysis and error diagnosis capabilities, significantly enhancing its autonomy and making it the first LLM-based AHD framework demonstrated to be effective on complex, ill-defined real-world problems.
\end{enumerate}

\section{Literature Review}

\subsection{Heuristic Design with Large Language Models}
The integration of LLMs with evolutionary computation now defines the frontier of Automatic Heuristic Design \cite{huang2025calm}. This approach has yielded a new generation of architectures capable of generating remarkably high-quality heuristics. A seminal example is FunSearch  \cite{romera2024mathematical}, which pairs a pre-trained LLM with a systematic evaluator to discover genuinely novel constructions for long-standing mathematical challenges, including the cap set and online bin packing problems. Building on this direction, the Evolution of Heuristics (EoH) \cite{liu2024evolution} architecture established a paradigm for generalizable and efficient design through the dual evolution of \texttt{thoughts} and code, supported by multi-strategy prompt engineering. The success of this approach has already inspired further development in multi-objective contexts, such as MEoH \cite{yao2025multi}. Concurrently, Reflective Evolution (ReEvo) \cite{ye2024reevo} introduced an innovative self-refining mechanism that substantially reduces the dependence on predefined heuristic component libraries, a traditional requirement of AHD. 

While powerful, these state-of-the-art architectures share a common paradigm rooted in natural selection: they apply evolutionary pressure directly to the generated heuristics. In this model, the LLM functions as a sophisticated \texttt{evolutionary operator} that produces candidate solutions, which are then evaluated and selected based on their performance. This focus on the final product, rather than the generative process that created it, represents a fundamental limitation that leaves significant potential for more intelligent and adaptive search untapped.

\subsection{Prompt Evolution for LLM-Driven Optimization}
Given the limitations of evolving heuristics, the logical next step is to evolve the generative process itself. A promising paradigm for this task is \textbf{Prompt Evolution} \cite{sahoo2024systematic}. Research in this area has demonstrated that iteratively refining an LLM's instructional prompt can dramatically improve the quality and specificity of its output. For instance, Li et al. (2023) designed a black-box evolutionary algorithm called SPELL that uses an LLM to iteratively improve its prompts \cite{li2023spell}. Similarly, the Promptbreeder mechanism was proposed to evolve and adapt prompts for specific domains by treating them as mutable tasks \cite{fernando2023promptbreeder}. Other works have confirmed the efficacy of using evolutionary algorithms to generate prompts through dynamic interaction with the model \cite{martins2023towards}.

These methods confirm that the reasoning pathway of an LLM is a viable and potent target for optimization. However, the application of prompt evolution to the structured, high-stakes domain of AHD remains a critical and unexplored research gap. In its current form, prompt evolution is often an unguided search, lacking a higher-level intelligence to direct its path efficiently and strategically. This raises a crucial question: how can a system learn to evolve its prompts in a principled manner?

\begin{figure*}[t]
    \centering % 图片居中
    \includegraphics[width=\textwidth]{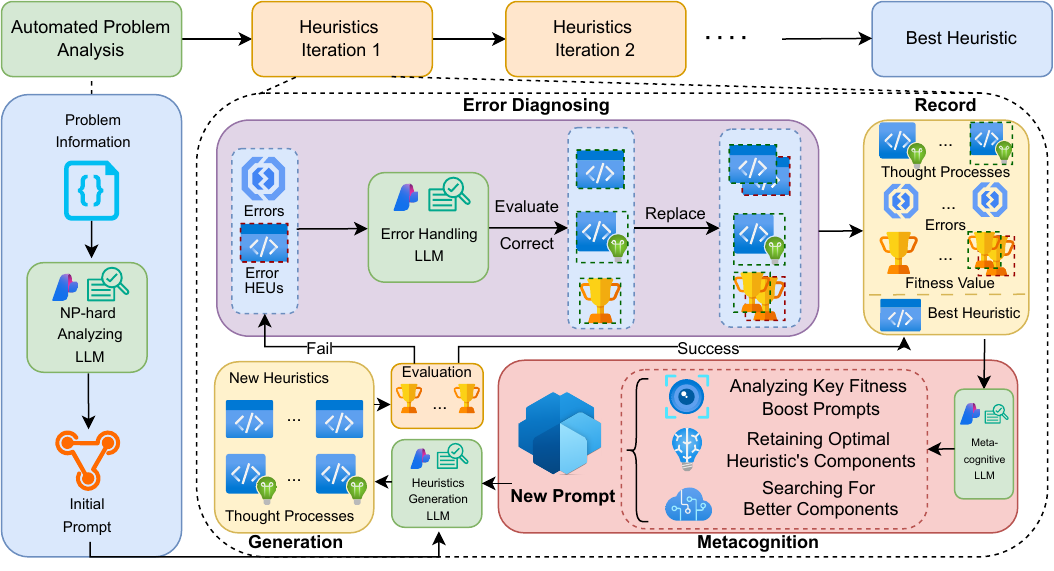} % 图片路径和宽度设置
    \caption{Flowchart of MeLA.}
    \label{fig:flowchart} % 图片标签，用于交叉引用
\end{figure*}

\subsection{Metacognition as a Framework for Self-Regulation}
A robust answer to this question is found in the cognitive science of \textbf{metacognition}. A fundamental distinction exists between cognition and metacognition: the cognitive system executes tasks (e.g., information retrieval), whereas the metacognitive system regulates these activities through higher-order processes of planning, monitoring, and evaluation \cite{lai2011metacognition}. Empirical studies demonstrate that individuals with well-developed metacognitive competence hold significant advantages: they can detect deficits in their own understanding and dynamically adjust their cognitive strategies based on performance outcomes \cite{akturk2011literature}. This capacity for self-regulation, often termed \texttt{learning to learn}, is crucial for adaptability in complex environments \cite{dunlosky2008metacognition,cox2005metacognition}.

This concept has recently found traction in AI research, with studies exploring how to instill metacognitive capabilities in AI agents \cite{lin2025think,wang2023metacognitive} and LLMs specifically \cite{li2025adaptive}. For heuristic design, metacognition offers the perfect blueprint for an architecture that does not just generate solutions, but actively analyzes its own problem-solving process to become a better problem-solver over time. However, despite its clear relevance, the strategic framework of metacognition has not yet been leveraged to guide the generative process within AHD.

\subsection{Summary and Identified Research Gap}
The foregoing review reveals a compelling opportunity at the intersection of three distinct fields. Current AHD architectures are powerful yet constrained by a paradigm that evolves heuristics. Prompt evolution offers a mechanism for process refinement but lacks strategic guidance. Finally, metacognition provides the theoretical blueprint for intelligent, autonomous self-regulation. This paper bridges these domains by introducing MeLA, an architecture that operationalizes metacognitive principles to drive prompt evolution, thereby forging a new and more powerful path for Automatic Heuristic Design.

\section{Methodology}
\subsection{Main Idea \& System Role}
MeLA introduces a paradigm shift in AHD by focusing on the metacognitive process of prompt evolution. It also incorporates automatic problem analysis and error correction mechanisms to enhance the architecture's versatility and robustness across a range of challenges, from classical benchmarks to complex real-world problems.

Previous architectures have concentrated on the generated heuristics themselves, employing crossover and mutation based on performance. This reliance on a natural selection-based evolutionary model can lead to instability, with inconsistent performance and significant variance among the generated heuristics. Furthermore, these approaches often lack the necessary mechanisms to effectively address the nuances of real-world problems and the errors that can arise in the generated heuristics.

In contrast, MeLA emphasizes the "\texttt{why such heuristics are generated}". This means MeLA focuses on the thought processes that LLMs generate when creating heuristics. The quality of this thought process directly correlates with the heuristic's performance; a sound line of reasoning leads to superior results, while flawed logic results in poor performance. By integrating metacognition, MeLA analyzes these LLM-generated thought processes to identify and retain the most effective reasoning patterns that contribute to improved heuristic performance. This allows the system to identify optimal heuristics and formulate new, more effective prompts to guide the LLM in subsequent heuristic generation, constituting a form of prompt evolution. The robustness of MeLA is further enhanced by its automated problem analysis and error diagnosis capabilities.

To operationalize this approach, MeLA introduces three distinct system roles:

\begin{itemize}
    \item \textbf{NP-hard Problem Analysis Expert:} This role is responsible for analyzing the code-based representations of NP-hard problems. It identifies key parameter characteristics and examines the fundamental challenges inherent in these problems, providing a solid foundation for heuristic design.

    \item \textbf{Elite Code Debugger:} This system component addresses errors within the generated heuristics. It analyzes execution failures, identifies the root causes of errors in the code, and applies targeted corrections while preserving the original problem-solving intent of the heuristic.

    \item \textbf{Metacognitive Reflector:} This role facilitates introspection on the thought processes and errors that occur during heuristic generation. Interpreted by the LLM as "another version of oneself", this component is crucial for self-regulation and strategic improvement.
\end{itemize}

\indent The detailed implementation of these system roles is further elaborated in the \textbf{Appendix}. A flowchart of the MeLA architecture is provided in \textbf{Figure 2}.

\begin{algorithm}
\caption{Iterative Heuristic Optimization}
\begin{algorithmic}[1]
\State \textbf{Input:} Problem $P$, Number of Initializations $N$, Generation Limit $L$
\State \textbf{Output:} Optimal Heuristic $Best(H)$

\Procedure{OptimizeHeuristic}{$P, L$}
    \For{$i \gets 1$ to $N$}
        \State $H, Th \gets \text{InitializeHeuristics}(\text{Analyze}(P))$
    \EndFor
    \For{$j \gets N+1$ to $L$}
        
        \State $f(H), E \gets \text{Evaluate}(H)$
        
        \If{$E \neq \emptyset$}
            \State $H, Th \gets \text{CorrectErrors}(H, E, Th)$
            \State $f(H) \gets \text{Evaluate}(H)$
        \EndIf
        
        \State $Meta \gets \text{AnalyzePerformance}(Th, f(H), $\\\hspace{5.75cm} $E,Best(H))$
        \State $H, Th \gets \text{GenerateNext}(Meta)$
    \EndFor
    
    \State \Return $Best(H)$
\EndProcedure
\end{algorithmic}
\end{algorithm}

\subsection{Iteration Steps}

The iterative optimization process of MeLA is detailed in Algorithm 1. At each iteration, MeLA maintains a population of heuristics, denoted as \( H = \{h_1, \dots, h_N\} \). Each heuristic \( h_i \) in this population is evaluated against a set of problem instances to determine its fitness value, \( f(h_i) \).

\indent The optimization loop begins with an initial problem \textbf{Analyze}, which informs the \textbf{InitializeHeuristics} step, where the first population of heuristics \(H\) and their corresponding thought processes \(Th\) are generated. Subsequently, the core iterative cycle commences. First, all heuristics in the current population are subjected to \textbf{Evaluation}, which yields their fitness scores \(f(H)\) and records any execution errors \(E\). If \(E\) are detected, the \textbf{CorrectErrors} procedure is invoked to repair the faulty heuristics and update the population \(H\), thought processes \(Th\) and fitness \(f(H)\). Following this, the \textbf{AnalyzePerformance} stage initiates a metacognitive review \(Meta\). It assesses the recorded thought processes \(Th\), the fitness landscape of the population \(f(H)\), the errors \(E\) and the characteristics of the best-performing heuristic \(Best(H)\) to generate a new set of refined prompts. These prompts are then used by the \textbf{GenerateNext} function to produce the next generation of heuristics \(H\) and thoughts \(Th\). This cycle of evaluation, correction, analysis, and generation repeats, driving the continuous refinement of the underlying problem-solving strategies.

\subsection{Prompt Evolution}

Prompt evolution in MeLA is a structured, reflective process driven by the metacognition phase. During this phase, MeLA leverages a comprehensive record of the iterative history—including the thought processes, execution errors, and resulting fitness values of all generated heuristics. The metacognitive analysis then evolves the guiding prompt by focusing on three key objectives:

\begin{itemize}
    \item \textbf{Reinforcement:} Identifying and reinforcing the reasoning patterns within the thought processes that correlated with fitness improvements.
    \item \textbf{Preservation:} Isolating and preserving high-performing components from the best heuristic, such as specific search strategies.
    \item \textbf{Innovation:} Hypothesizing novel components or modifications that could lead to further performance gains.
\end{itemize}

\indent A concrete example illustrating this prompt evolution process is detailed in the \textbf{Appendix}.

\subsection{Challenges in Real-World Problems}

While architectures like EoH and ReEvo have demonstrated commendable performance on classical benchmark problems, they face significant challenges when applied to real-world optimization scenarios. These problems are often high-dimensional and richly parameterized, requiring meticulously detailed descriptions to guide the LLM toward generating feasible heuristics. Crafting such descriptions is non-trivial; it demands deep user expertise both in the problem domain and in the nuances of interacting with LLMs. This dependency on manual, expert-driven problem formulation limits their applicability and scalability for diverse real-world challenges. For a detailed guide on the modifications required to enable the execution of EoH and ReEvo on these complex problems, please refer to the \textbf{Appendix}.

\indent Compounding this issue, even with well-crafted problem descriptions, both EoH and ReEvo exhibit a high probability of generating non-executable heuristics. This stems from the inherent difficulty an LLM faces in translating complex, multi-parameter problem specifications into syntactically and logically correct code, leading to errors such as undefined parameters or out-of-bounds array access. While this has a lesser impact on an architecture like EoH, which does not strictly depend on a population of valid individuals, it poses a critical failure point for ReEvo. The crossover mechanism in ReEvo requires executable parent heuristics; if none can be successfully generated, the evolutionary process halts entirely.

\indent MeLA is specifically engineered to overcome these limitations. It utilizes an LLM's analytical capabilities to approach problem understanding from an expert perspective, thereby enhancing the accuracy and versatility of heuristic generation. Crucially, it integrates a robust error diagnosis and correction mechanism to dramatically increase the rate of producing executable code.

\subsection{Predefined Prompts}

MeLA's operation is guided by four distinct, predefined prompts designed to steer the LLM through the heuristic design process.

\begin{itemize}
    \item \textbf{Problem Analysis Prompt:} The LLM first analyzes the problem's complete Python code to produce an expert description of its key characteristics and inherent computational difficulty. This analysis grounds the entire process.

    \item \textbf{Generation Prompts:} These prompts create the heuristics. The \textbf{Initial Generation} prompt uses the problem analysis to produce a diverse population of \(N\) heuristics. Subsequently, the \textbf{Metacognitive Generation} prompt uses insights from the metacognition step to evolve this population, generating \(N\) refined heuristics for the next iteration.

    \item \textbf{Error Prompt:} If a heuristic fails during evaluation, the LLM receives the faulty code and its corresponding error message. It then attempts to generate a corrected version up to \(M\) times. The best-performing valid candidate replaces the original; if all attempts fail, the heuristic's failure is logged.

    \item \textbf{Metacognition Prompt:} This core prompt drives prompt evolution by instructing the LLM to analyze the iteration's complete history (all thought processes, errors, and fitness values). It performs a self-reflective analysis to: 1) identify reasoning patterns that led to improved fitness, 2) preserve high-performing components (e.g., a search strategy like \texttt{L\'evy Flight}) from the best heuristic, and 3) hypothesize new strategies for further enhancement. The output of this prompt directly informs the Metacognitive Generation prompt.
\end{itemize}

\indent Full prompt details are provided in the \textbf{Appendix}.

\begin{table*}[htbp]
\centering
\small
\begin{tabular}{lcccccccc}
\toprule
\multirow{2}{*}{Method} & \multicolumn{2}{c}{TSP50} & \multicolumn{2}{c}{BPP500} & \multicolumn{2}{c}{ACS} & \multicolumn{2}{c}{WSN} \\
\cmidrule(lr){2-3} \cmidrule(lr){4-5} \cmidrule(lr){6-7} \cmidrule(lr){8-9}
 & SR ($\uparrow$) & Obj. ($\downarrow$) & SR ($\uparrow$) & Obj. ($\downarrow$) & SR ($\uparrow$) & Obj. ($\downarrow$) & SR ($\uparrow$) & Obj. ($\downarrow$) \\
\midrule
GA & -- & $16.75 \pm 0.72$ & -- & $219.60 \pm 0.45$ & -- & $3927.10 \pm 976.76$ & -- & $1923.17 \pm 0.34$ \\
PSO & -- & $11.46 \pm 0.34$ & -- & $219.30 \pm 0.46$ & -- & $4067.25 \pm 1095.79$ & -- & $2674.44 \pm 326.81$ \\
SCSO & -- & $7.87 \pm 0.24$ & -- & $221.71 \pm 0.39$ & -- & $3875.83 \pm 1939.60$ & -- & $661.00 \pm 199.52$ \\
SOA & -- & $20.48 \pm 0.39$ & -- & $220.22 \pm 0.28$ & -- & $6349.44 \pm 1504.71$ & -- & $943.69 \pm 80.90$ \\
WO & -- & $20.97 \pm 0.29$ & -- & $221.69 \pm 0.37$ & -- & $6975.08 \pm 7256.24$ & -- & $1251.25 \pm 670.21$ \\
\midrule
EoH$^*$ & 70.97\% & $5.90 \pm 0.02$ & 88.54\% & $209.08 \pm 4.77$ & 75.49\% & $1252.88 \pm 794.44$ & 53.36\% & $177.84 \pm 96.32$ \\
ReEvo$^*$ & 89.46\% & $5.89 \pm 0.03$ & \textbf{96.73\%} & $208.76 \pm 2.06$ & 40.71\% & $1848.78 \pm 406.90$ & 56.52\% & $117.61 \pm 18.30$ \\
\midrule
\textbf{MeLA} & \textbf{98.41\%} & $\mathbf{5.85 \pm 0.01}$ & 93.28\% & $\mathbf{205.48 \pm 0.84}$ & \textbf{94.07\%} & $\mathbf{589.60 \pm 22.79}$ & \textbf{99.21\%} & $\mathbf{53.89 \pm 2.77}$ \\
\bottomrule
\end{tabular}
\caption{Comparative Performance Analysis Across Different Problems}
\label{tab:comparative}
\end{table*}

\begin{figure*}[t]
    \centering % 图片居中
    \includegraphics[width=\textwidth]{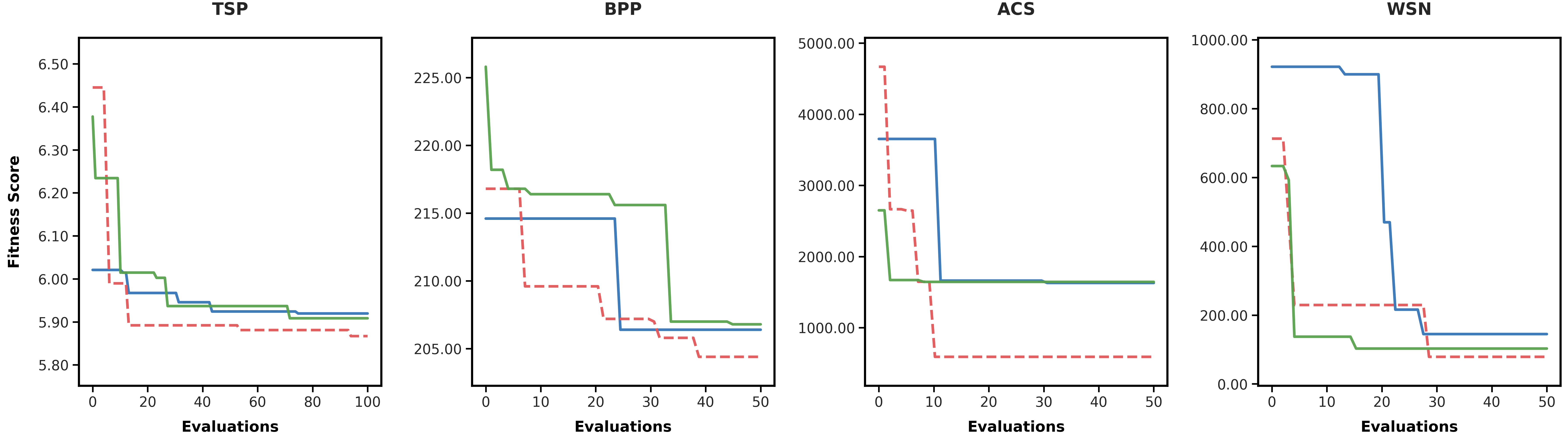} % 图片路径和宽度设置
    \caption{Fitness Values of Different Architectures on Different Problems. (
	 \protect\tikz[baseline]{\protect\draw[myred, line width=0.3mm,densely dashed] (0,.8ex)--++(0.5,0);}~MeLA,
	 \protect\tikz[baseline]{\protect\draw[myblue, line width=0.3mm] (0,.8ex)--++(0.5,0) ;}~ EoH, 
     \protect\tikz[baseline]{\protect\draw[mygreen, line width=0.3mm] (0,.8ex)--++(0.5,0) ;}~ ReEvo
     )}
    \label{fig:fitness} % 图片标签，用于交叉引用
\end{figure*}

\section{Experiments}
\subsection{Experimental Setup}

To rigorously evaluate the performance and versatility of MeLA, we designed an experimental suite comprising three distinct categories of optimization challenges: a classical benchmark, a black-box benchmark, and two complex real-world problems. This selection allows for a comprehensive assessment of the architecture's capabilities across various scenarios. The chosen problems are:

\begin{itemize}
    \item \textbf{Traveling Salesperson Problem (TSP):} A canonical NP-hard problem requiring the determination of the most efficient route that visits a set of cities exactly once before returning to the origin. In the experiment, the initial population size is 30, the evolutionary population size is 10, and a total of 100 solutions are generated. 
    
    \item \textbf{Bin Packing Problem (BPP):} A classic black-box combinatorial optimization problem that involves packing items of varying sizes into the minimum number of fixed-capacity containers \cite{ross2002hyper}. It serves as a benchmark for resource allocation challenges. In the experiment, the initial population size is 30, the evolutionary population size is 10, and a total of 50 solutions are generated.   
    
    \item \textbf{Adaptive Curriculum Sequencing (ACS):} A complex, real-world combinatorial optimization problem from the domain of e-learning. ACS involves generating personalized learning paths by balancing learner characteristics (e.g., knowledge level, attention span) with pedagogical constraints (e.g., concept dependencies, difficulty levels). As an NP-Hard problem, it requires optimizing multiple, often conflicting, objectives. In the experiment, the initial population size is 20, the evolutionary population size is 10, and a total of 50 solutions are generated.
    
    \item \textbf{Wireless Sensor Network (WSN) Deployment:} A real-world optimization problem from the communications domain, abbreviated herein as WSN. The goal is to optimize the placement and power management of sensor nodes to maximize network coverage and connectivity while minimizing total energy consumption, addressing NP-hard challenges like the Minimum Power k-Coverage problem. In the experiment, the initial population size is 20, the evolutionary population size is 10, and a total of 50 solutions are generated.
\end{itemize}

 In TSP and BPP problems, we optimize the heuristic information based on Ant Colony Optimization (ACO). The core mechanism of ACO involves guiding random solution sampling through heuristic measures. This mechanism is naturally suitable for validating the value of AHD, which automatically designs these heuristic rules to guide the search process, replacing manually designed heuristic functions\cite{ye2024reevo}. In ACS and WSN problems, we optimize the heuristic algorithm. By evaluating the generated results of heuristic information and heuristic algorithms, we assess the excellent performance of MeLA.

To ensure experimental fairness, EoH, ReEvo, and MeLA all use the same experimental parameters.

The selection of ACS and WSN was specifically intended to test MeLA's cross-domain execution capabilities, and all four problems are formulated as minimization tasks. We selected DeepSeek-V3-0324 as the core generative engine due to its strong metacognitive reasoning capabilities. As our preliminary tests confirmed that performance is primarily driven by the MeLA architecture itself, not the specific LLM, this work focuses on validating the framework with a single capable model. 

For our comparative analysis, we benchmarked MeLA against five traditional metaheuristics (Genetic Algorithm (GA) \cite{holland1992genetic}, Particle Swarm Optimization (PSO) \cite{kennedy1995particle}, Sand Cat Swarm Optimization (SCSO) \cite{seyyedabbasi2023sand},  Skill Optimization Algorithm (SOA) \cite{givi2023skill}, and Walrus Optimization (WO) \cite{han2024walrus}) and two state-of-the-art LLM-driven architectures (Evolution of Heuristics (EoH) \cite{liu2024evolution} and Reflective Evolution (ReEvo) \cite{ye2024reevo}).

It should be emphasized that for excellent algorithms such as GA, in the TSP and BPP problems, we evaluate the ACO's performance with heuristic information designed by these algorithms, while in the ACS and BPP problems, we evaluate the algorithms themselves.

The details of these comprehensive definitions and constraints for all problems are documented in the \textbf{Appendix}.

\begin{figure*}[t]
    \centering % 图片居中
    \includegraphics[width=\textwidth]{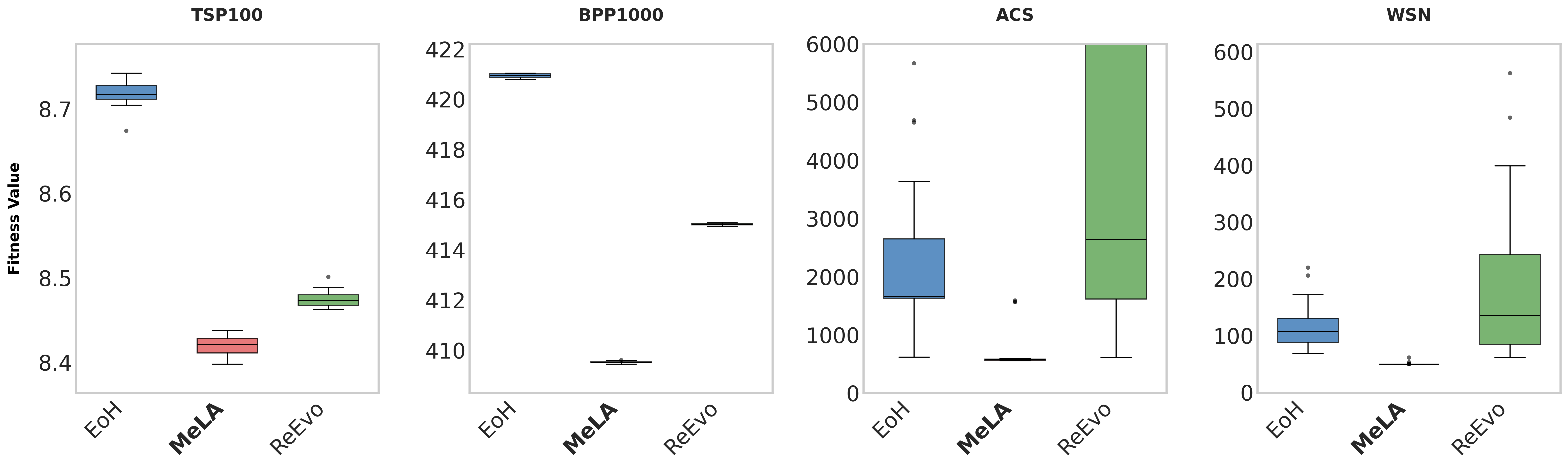} % 图片路径和宽度设置
    \caption{Performance of Optimal Heuristic Generated by Different Architectures.}
    \label{fig:sta} % 图片标签，用于交叉引用
\end{figure*}

\subsection{Results and Analysis}

The experimental results across the four problem domains are summarized in \textbf{Figure 3} and \textbf{Table 1}. \textbf{Figure 3} illustrates a representative convergence plot from one of our independent runs. Due to the stochastic nature of LLM-based generation, an optimal heuristic may occasionally be found during initialization; presenting a run that demonstrates clear iterative improvement provides a more insightful view of an architecture's dynamic behavior. As shown, MeLA consistently achieves superior fitness values across all four problems, with particularly significant advantages over EoH and ReEvo on the BPP and ACS problems.

A quantitative summary of our findings over three independent runs, including heuristic success rates (SR) and final fitness values, is presented in \textbf{Table 1}. The data confirms MeLA's superior performance on the TSP, ACS, and WSN problems in terms of both success rate and average final fitness. The advantage is most pronounced in real-world scenarios. For the ACS problem, MeLA's success rate is 18.58\% and 53.36\% higher than EoH and ReEvo, respectively, while achieving a final fitness that is 52.94\% and 68.11\% better. Similarly, on the WSN problem, MeLA achieves a 69.70\% and 54.18\% higher success rate and improves the final fitness by 71.30\% and 48.27\%. For the BPP problem, while ReEvo records a marginally higher success rate (by 3.45\%), MeLA still secures the best average fitness, indicating its overall effectiveness. Finally, when compared against traditional metaheuristics on the ACS and WSN problems, all modern AHD architectures demonstrate a clear advantage, affirming the power of this paradigm for complex, multi-parameter optimization.

Beyond aggregate performance, we analyzed the generality and stability of the single best heuristic generated by each architecture, with results detailed in \textbf{Figure 4}. For the TSP and BPP, we tested generalization by evaluating the heuristics on 64 different larger-scale instances over 10 independent runs. For the complex ACS and WSN problems, we tested performance stability by re-executing the optimal heuristics 30 times with different random seeds. In all scenarios, the heuristic generated by MeLA demonstrated the most stable and superior performance. This was particularly evident for the ACS and WSN problems, where the heuristic not only exhibited minimal deviation but consistently converged to its optimal fitness value in nearly every run, confirming its exceptional quality and reliability.

The specific parameters used for all problems and the code for the optimal heuristics generated by each architecture are available in the \textbf{Appendix}.

\subsection{Ablation Analysis}

To validate the individual contributions of MeLA's core components, we conducted a series of ablation studies. We systematically analyzed the impact of: Prompt Evolution, the Automated Problem Analysis mechanism, the Error Diagnosis mechanism, and the Metacognitive Search component.

\begin{itemize}
    \item \textbf{Prompt Evolution:} The superiority of Prompt Evolution over traditional natural selection is demonstrated by MeLA's overall performance in \textbf{Table 1} and \textbf{Figure 4}. On classic benchmarks, it yields heuristics with superior fitness, an advantage particularly pronounced on the BPP. On real-world problems, it produces solutions that are not only higher-performing but also significantly more stable. This confirms that evolving the underlying reasoning process is a more effective and robust strategy than evolving the heuristic code directly.

    \item \textbf{Automated Problem Analysis (PA) Mechanism:} To isolate the effect of the PA mechanism, we compared initialization performance with and without this component. As shown in \textbf{Table 2}, the PA mechanism provides a moderate fitness improvement across problems and, critically, yields a significant increase in the initial heuristic success rate on the black-box BPP challenge.

    \item \textbf{Error Diagnosis Mechanism:} The efficacy of the Error Diagnosis mechanism is clearly evidenced by the heuristic success rates presented in \textbf{Table 1}. MeLA's ability to achieve near-perfect execution rates, especially on complex real-world problems where baselines like EoH and ReEvo falter, is a direct result of this component's ability to autonomously repair faulty code.

    \item \textbf{Metacognitive Search Component:} The impact of the Metacognitive Search component is demonstrated by comparing the average fitness values before and after its application (\textbf{Table 2}). For the ACS problem, each successive metacognitive stage delivers a clear and consistent improvement in fitness. On the BPP, the first metacognitive stage provides a substantial fitness gain, with subsequent stages showing diminishing returns as the heuristics approach a strong optimum. Furthermore, this component helps maintain or improve heuristic success rates throughout the evolutionary process, as seen in the final 100\% success rate for ACS after the third stage.
\end{itemize}

\begin{table}[htbp]
\centering
\small
\begin{tabular}{lcccccc}
\toprule
\multirow{2}{*}{PA/Meta} & \multicolumn{2}{c}{BPP} & \multicolumn{2}{c}{ACS} \\
\cmidrule(lr){2-3} \cmidrule(lr){4-5}
 & Avg. ($\downarrow$) & SR($\uparrow$) & Avg. ($\downarrow$) & SR($\uparrow$) \\
\midrule
Have PA & $221.50$ & 70.00\% & $4434.14$ & 100.00\% \\
No PA & $221.80$ & 56.67\% & $4447.32$ & 93.33\%  \\
\midrule
Initial & $219.61$ & 86.67\% & $5335.47$ & 95.00\% \\
Meta-1 & $207.16$ & 90.00\% & $4311.72$ & 90.00\% \\
Meta-2 & $207.20$ & 90.00\% & $3642.88$ & 90.00\% \\
Meta-3 & -- & -- & $3530.78$ & 100\% \\
\bottomrule
\end{tabular}
\caption{Ablation results.}
\label{tab:combined_results}
\end{table}

\section{Conclusion}
This paper introduces MeLA, a novel architecture that marks a significant advance in Automatic Heuristic Design. By pioneering \textit{prompt evolution}, MeLA shifts the focus from evolving heuristic code—the standard in natural selection-based methods—to evolving the underlying reasoning process that generates the heursitic. Central to our approach is a metacognitive framework that enables the system to iteratively refine its generative strategy. This allows MeLA to discover high-performing thought patterns to get superior solution quality and execution success rates across a diverse range of optimization challenges.

This shift from direct code manipulation to cognitive framework optimization allows MeLA to overcome the limitations of prior approaches. Furthermore, by making the thought process the target of optimization, MeLA yields a unique and powerful form of interpretability. Unlike evolving cryptic heuristics, MeLA produces an optimized prompt, which is human-readable. This prompt serves as a reusable, high-level strategy that can be understood, manually refined, and easily adapted by human experts, bridging the gap between automated discovery and human-centric problem-solving. Our results validate prompt evolution as a powerful and robust paradigm, offering a more stable and adaptable method for generating high-quality heuristics, particularly for complex real-world problems.

\indent For future work, a key direction will be to investigate the performance of MeLA with different underlying LLMs. Such experiments will not only explore the trade-offs between various models but also further validate the model-agnostic nature of the MeLA framework, enhancing its generalizability.

\section*{Acknowledgments}
This work is supported by the Ningbo Municipal Bureau of Science and Technology (Grant No. 2025Z197).

\bibliography{reference}

\appendix
\setcounter{figure}{0}
\setcounter{table}{0}
\renewcommand{\thefigure}{A\arabic{figure}}
\renewcommand{\thetable}{A\arabic{table}}

\setlength{\leftmargini}{20pt}
\makeatletter\def\@listi{\leftmargin\leftmargini \topsep .5em \parsep .5em \itemsep .5em}
\def\@listii{\leftmargin\leftmarginii \labelwidth\leftmarginii \advance\labelwidth-\labelsep \topsep .4em \parsep .4em \itemsep .4em}
\def\@listiii{\leftmargin\leftmarginiii \labelwidth\leftmarginiii \advance\labelwidth-\labelsep \topsep .4em \parsep .4em \itemsep .4em}\makeatother

\setcounter{secnumdepth}{0}
\renewcommand\thesubsection{\arabic{subsection}}
\renewcommand\labelenumi{\thesubsection.\arabic{enumi}}

\newcounter{checksubsection}
\newcounter{checkitem}[checksubsection]

\newcommand{\checksubsection}[1]{%
  \refstepcounter{checksubsection}%
  \paragraph{\arabic{checksubsection}. #1}%
  \setcounter{checkitem}{0}%
}

\newcommand{\checkitem}{%
  \refstepcounter{checkitem}%
  \item[\arabic{checksubsection}.\arabic{checkitem}.]%
}
\newcommand{\question}[2]{\normalcolor\checkitem #1 #2}
\newcommand{\ifyespoints}[1]{\makebox[0pt][l]{\hspace{-15pt}\normalcolor #1}}

\onecolumn
\section*{Appendix}

\subsection{A.1 Errors}

Here are the errors that may occur during the heuristic operation of LLM generation.

\begin{figure*}[htbp]
\begin{center}
    \begin{tcolorbox}[
        arc=3mm, % 圆角半径
        colback=yellow!5, % 背景色
        colframe=black, % 边框色
        boxrule=1pt, % 边框粗细
        width=\textwidth
    ]
        ValueError: operands could not be broadcast together with shapes (50,) (50,150)\\
TypeError: 'numpy.float64' object is not callable\\
IndexError: too many indices for array: array is 1-dimensional, but 2 were indexed\\
ValueError: operands could not be broadcast together with shapes (50,) (50,150)\\
IndexError: invalid index to scalar variable.\\
ValueError: operands could not be broadcast together with shapes (50,150) (50,)\\
SyntaxError: '[' was never closed\\
SyntaxError: Generator expression must be parenthesized\\
SyntaxError: '[' was never closed\\
SyntaxError: '[' was never closed\\
ValueError: probabilities contain NaN\\
SyntaxError: closing parenthesis ']' does not match opening parenthesis '('\\
SyntaxError: '[' was never closed
    \end{tcolorbox}
    \caption{Incomplete problem descriptions will result in no runnable heuristics for both EoH and ReEvo.}
\end{center}
\end{figure*}

\subsection*{A.2 Prompt Evolution}

\indent The process of prompt evolution based on metacognition is shown in the figure below. As illustrated in the figure, this flowchart demonstrates a systematic metacognition-driven prompt evolution framework integrating thought processes, fitness evaluation, and large language model (LLM) interaction. The left panel displays initial heuristic configurations (v1) with corresponding fitness scores (3656.5 and 1669.8125), featuring optimization components such as exploration/exploitation balance and adaptive cosine-based movement. The central metacognition module performs critical analysis, evaluating key aspects including the retention of Levy flights for global search and dynamic scaling mechanisms. This reflective process informs the LLM's prompt generation, yielding an optimized thought process (heuristics\_v2) on the right, which achieves significantly improved performance (fitness: 615.0625) through enhanced Levy flight implementation and fitness-aware restart mechanisms. The diagram visually captures the iterative refinement cycle where metacognitive analysis guides LLM-mediated prompt optimization, demonstrating measurable fitness improvement across generations. An example can be seen in \textbf{Figure A2}.

\begin{figure*}[htbp]
    \centering % 图片居中
    \includegraphics[width=\textwidth]{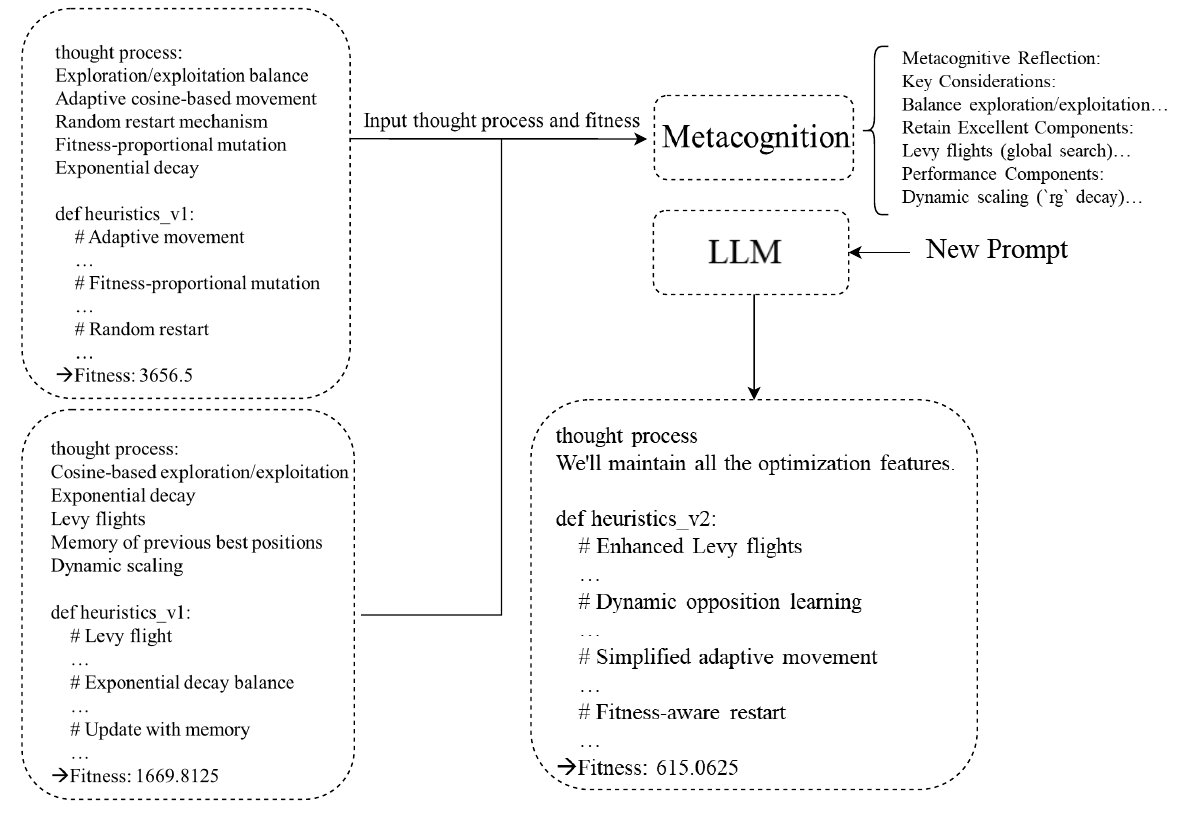} % 图片路径和宽度设置
    \caption{An example of Prompt Evolution.}
    \label{fig:fitness} % 图片标签，用于交叉引用
\end{figure*}

\subsection{A.3 The Problem Descriptions of EoH and ReEvo}

\indent The problem description for EoH and ReEvo needs to be designed as follows for proper operation. Incomplete problem descriptions will result in no runnable heuristics for both EoH and ReEvo. For EoH, with a complete problem description, it achieves a runnable heuristic generation rate of 75.49\%, while ReEvo only reaches 40.41\%. However, ReEvo can now run normally under these conditions. Refer to \textbf{Figure A3} for details.

\begin{figure*}[htbp]
\begin{center}
    \begin{tcolorbox}[
        arc=3mm, % 圆角半径
        colback=yellow!5, % 背景色
        colframe=black, % 边框色
        boxrule=1pt, % 边框粗细
        width=\textwidth
    ]
        \textbf{Initial Description}\\
        Implement a function that performs adaptive course material recommendation using the SCSO metaheuristic optimization algorithm.

        \textcolor{green!70!black}{Heuristic operation success rate}
        : EOH - 5\% ReEvo - The operation rate is too low to proceed to the crossover and mutation stages.
        
        \textbf{Complete Description}\\         
        Implement a function that performs adaptive course material recommendation using the SCSO metaheuristic optimization algorithm. The function must:\\
        1. Strict Requirements for Parameter Access:\\
   - All parameters must be accessed using OBJECT.ATTRIBUTE notation only\\
   - NEVER use dictionary-style access like data\_al['lb'] or data\_al[0]\\
   - Required parameter accesses that must use dot notation:\\
   * data\_al.lb - lower bounds\\
     * data\_al.ub - upper bounds\\
     * data\_al.dim - problem dimension  \\
     * data\_al.SearchAgents - population size\\
     * data\_al.MaxIter - maximum iterations\\
     2. Mandatory Implementation Structure:\\
   \# 1. Parameter initialization (MUST use dot notation)\\
   lb = np.array(data\_al.lb)\\
   ub = np.array(data\_al.ub)\\
   dim = data\_al.dim\\
   SearchAgents\_no = data\_al.SearchAgents\\
   \# 2. Position initialization \\
   ub\_array = np.array(ub)\\
   lb\_array = np.array(lb)\\
   position initialization logic\\
   3. Core Algorithm Requirements:\\
   - Must implement both exploration and exploitation phases\\
   - Must include boundary constraint handling\\
   - Must use cosine-based position updates\\
   - Must maintain roulette wheel selection\\
4. Input/Output Specifications:\\
   Input Parameters:\\
   - data\_al: Algorithm config object with dot-accessible attributes\\
   - data\_pb: Problem data object\\
   - Positions: Current population positions\\
   - Best\_pos: Current best solution\\  
   - Best\_score: Current best fitness\\
   - rg: Current search radius\\
   Returns:\\
   - Updated Positions array only\\
   - NO other return values allowed\\
       \textcolor{green!70!black}{Heuristic operation success rate}
            : EOH - 75.49\% ReEvo - 40.71\%
    \end{tcolorbox}
    \caption{Incomplete problem descriptions will result in no runnable heuristics for both EoH and ReEvo.}
\end{center}
\end{figure*}

\subsection{A.4 System Role \& Full Prompt}

For details on the specific system roles and various prompt expressions, please refer to \textbf{Figure A4} and \textbf{Figure A5}.

\begin{figure}
\begin{center}
    \begin{tcolorbox}[
        arc=3mm, % 圆角半径
        colback=blue!5, % 背景色
        colframe=black, % 边框色
        boxrule=1pt, % 边框粗细
        width=\textwidth
    ]
        \textbf{NP-hard Problem Analysis Expert}\\
        Act as an NP-hard problem analyst.\\ 
Please analyze the characteristics and solutions of the NP-hard problem.\\
        \textbf{Elite Code Debugger}\\
        You are an elite Code Debugging.\\
Please correct the Python code.\\
The code generation format must strictly follow the example below:\\
    \{thought process:\\
    1.xxx\\
    2.xxx\\
    ...\}\\   
    ```python\\
    import numpy as np \\
    def heuristics\_v1(data\_al, data\_pb, Positions, Best\_pos, Best\_score, rg):\\
        * The rest remains unchanged. *\\
        \#EVOLVE-START\\
        * Your optimized code *\\
        \#EVOLVE-END   \\    
        return Positions\\
    ```  
    \textbf{Metacognition Role}\\
    You are another version of yourself, a process of thinking and a set of errors through which you come to understand yourself.\\
Please analyze the thought process records, errors, and the optimal algorithm. 
    
    \end{tcolorbox}
    \vspace{-10pt}
    \caption{System Role}
\end{center}

\begin{center}
    \begin{tcolorbox}[
        arc=3mm, % 圆角半径
        colback=blue!5, % 背景色
        colframe=black, % 边框色
        boxrule=1pt, % 边框粗细
        width=\textwidth
    ]
        \textbf{Problem Analysis Prompt}\\
I have an existing problem with the code as follows:\{problem\}.\\
Please analyze the problem.\\
Your analysis must be within 50 words.\\
        \textbf{Generation Prompt}\\
        \textbf{Generation 1}\\
        \{problem\}\\
I have an initial algorithm with the code as follows:
\{init\_code\}.\\
And the optimization history it presents in the problem is as follows:\{init\_eval\}.\\
Please help me create a new algorithm that has a totally different form from the given ones but can be motivated from them.\\
1.Analyze the history of fitness values and optimize the algorithm with the goal of surpassing the optimal value.\\
2.You will notice that there are \#EVOLVE-START and \#EVOLVE-END comments in the following code. The code within these comments is the part that you need to optimize.\\
3.The code:\{code\}. Analyze the algorithm, optimize the algorithm. Record your thought process in the \{\} brackets. \\
4.Your thought process must be within 50 words.\\
    \textbf{Generation 2}\\
    The reflection results of metacognition are as follows:\{metacognition\}.\\
I have a existing algorithm with their codes as follows:\{init\_code\}.\\
The optimization history it presents in the optimization problem is as follows:\{init\_eval\}.\\
Please retain the advantageous components and innovate to improve the deficient ones.\\
1.You will notice that there are \#EVOLVE-START and \#EVOLVE-END comments in the following code. The code within these comments is the part that you need to optimize.\\
2.The code:\{code\}. Analyze the algorithm, optimize the algorithm.Record your thought process in the \{\} brackets. \\
3.Your thought process must be within 50 words.\\
    \textbf{Error Prompt}\\
    The code you generated \{code\_str\} has the following error \{str\_error\}.\\
Please make it correct and functional.\\
    \textbf{Metacognition Prompt}\\
    The thinking process and score of each algorithm are as follows:\{thoughts\}.\\
The errors are as follows:\{errors\}.\\
You should avoid the errors and ensure that no new error.
The optimal algorithm is as follows:\{code\}.\\
Please conduct metacognitive reflection on your own thinking process, scores and mistakes.\\
1.Analyze the important considerations for optimizing fitness values.\\
2.The excellent components that should be retained in the optimal algorithm.\\
3.The components with better performance that need to be retained.\\
4.Your output content must be within 80 words.       
    \end{tcolorbox}
    \vspace{-10pt}
    \caption{Full Prompt}
\end{center}
\end{figure}

\subsection{A.5 Problems}

All experiments were performed on a 48-core computer with an Intel Xeon Gold 6248R 3.00GHz CPU, running Windows 10 Pro and equipped with 255GB of RAM.

The relevant parameters for the four questions are shown in Table 1. When experimenting with different problem sizes for TSP and BPP, we ensure the use of the same dataset, with its generation criteria based on ReEvo \cite{ye2024reevo}. This paper will not provide further explanation for these two experiments. If needed, please directly refer to the sections in ReEvo concerning TSP50, TSP100, BPP500, and BPP1000. The operating parameters for these two issues are shown in \textbf{Table A1} and \textbf{Table A2}.

\indent This paper will now focus on explaining the ACS and WSN problems.

\indent $\bullet$ ACS: In the ACS problem, we consider real-world scenarios and utilize the Open University Learning Analytics Dataset (OULAD) for modeling and optimizing ACS solutions. It includes data from 10,000 learners, information on 7 selected courses, learner profiles, and their interactions with the virtual learning environment.  To reflect realistic teaching conditions, we set the number of learners enrolled in a single course to 30, with a total of 20 concepts to be mastered in the course. The learning materials consist of 120 items, each covering at least one concept, but none covering all 20 concepts. Due to the high complexity of the problem, in evolutionary computation, we generally use multiple iterations of Search Agents in a single experiment to find feasible solutions. Therefore, in this problem, a heuristic evaluation method is generated by having 20 agents search for the optimal solution over 50 iterations to assess the performance of the heuristic. The relevant operating parameters are shown in \textbf{Table A3}. 

\indent The constraints in the ACS experimental setup are carefully designed to model realistic learning constraints and priorities. The penalty factor $\varepsilon_1$ (set to 1) applies when learners exceed the required number of concepts, while $\varepsilon_2$ ($10^4$) imposes a significantly higher penalty for failing to cover essential concepts, emphasizing comprehensive learning coverage. The attention span constraint is enforced through $\varepsilon_3$ (1000), penalizing excessive cognitive load. Weight coefficients $\omega_1$ (0.25), $\omega_2$ (1), and $\omega_3$ (0.25) balance these penalties, with $\omega_2$ giving strongest emphasis to concept coverage completeness. Priority-based material limits are implemented through $\psi_1$ (3) for high-priority content, $\psi_2$ (6) for medium-priority, and $\psi_3$ (1) for challenging materials, ensuring appropriate distribution of learning resources. These parameters collectively optimize the Adaptive Course Sequencing problem using OULAD data. Its constraint parameter table is detailed in \textbf{Table A5}. 

\indent $\bullet$ WSN: Wireless Sensor Networks deployment problem involving 200 fixed-position Sensing Nodes (SNs) and 50 Convergence Nodes (CNs) whose positions and transmission powers need to be optimized. The system aims to achieve three primary objectives: ensuring all SNs are connected to at least one CN, maintaining connectivity among all CNs, and minimizing total power consumption while meeting technical requirements. The network operates with CNs having a 20-unit communication range and a capacity constraint of serving no more than 15 SNs each, using a path loss model that includes a 55 dB base loss, 2.5 path loss exponent, additional quadrant-based loss factors ($\beta_x$, $\beta_y$), and a minimum required signal strength of -85 dBm. The optimization employs a population-based metaheuristic approach with 50 search agents running for 100 iterations, where each solution represents CN configurations through three parameters (x position, y position, and power level) bounded within [0,50] for coordinates and [0,30] dBm for power. The fitness function combines multiple components including coverage penalties (10$\times$ per uncovered SN), connectivity penalties (1000 for disconnected CN networks), power uniformity penalties (100$\times$ for excessive standard deviation beyond 1 dB), and the true fitness metric of total power consumption converted from dBm to milliwatts. A feasible solution must satisfy all constraints by covering all 200 SNs, maintaining full CN connectivity, keeping power uniformity within limits, and achieving a fitness score below the 1000 penalty threshold, with the framework providing convergence tracking and visualization capabilities to monitor optimization progress throughout this complex multi-constraint problem. For details of the parameters and constraints, please refer to \textbf{Table A4} and \textbf{Table A6}.

\indent The parameters of each meta-heuristic algorithm are shown in \textbf{Table A7}.

\begin{table*}[htbp]
\centering
\begin{tabular}{ll}
\toprule
\textbf{Parameter Category} & \textbf{TSP} \\
\midrule
LLM architecture & DeepSeek-V3-0324 \\
LLM Temperature & 1 \\
Population Size & 30 (initial), 10 (other) \\
Independent Runs & 3 \\
Solutions Generated & 100 \\
\bottomrule
\end{tabular}
\caption{Parameter settings for the TSP problem}
\vspace{0.5cm}

\begin{tabular}{ll}
\toprule
\textbf{Parameter Category} & \textbf{BPP} \\
\midrule
LLM architecture & DeepSeek-V3-0324 \\
LLM Temperature & 1 \\
Population Size & 30 (initial), 10 (other) \\
Independent Runs & 3 \\
Solutions Generated & 50 \\
\bottomrule
\end{tabular}
\caption{Parameter settings for the BPP problem}
\vspace{0.5cm}

\begin{tabular}{ll}
\toprule
\textbf{Parameter Category} & \textbf{ACS} \\
\midrule
LLM architecture & DeepSeek-V3-0324 \\
LLM Temperature & 1 \\
Population Size & 20 (initial), 10 (other) \\
Independent Runs & 3 \\
Solutions Generated & 50 \\
\addlinespace
\midrule
Problem Parameter & \\
\quad Materials & 120 \\
\quad Concepts & 20 \\
\quad Students & 30 \\
\addlinespace
\midrule
Iterative Parameter & \\
\quad Search Agents & 20 \\
\quad Max Iterations & 50 \\
\bottomrule
\end{tabular}
\caption{Parameter settings for the ACS problem}
\vspace{0.5cm}

\begin{tabular}{ll}
\toprule
\textbf{Parameter Category} & \textbf{WSNs} \\
\midrule
LLM architecture & DeepSeek-V3-0324 \\
LLM Temperature & 1 \\
Population Size & 20 (initial), 10 (other) \\
Independent Runs & 3 \\
Solutions Generated & 50 \\
\addlinespace
\midrule
Problem Parameter & \\
\quad Number of CN & 50 \\
\quad Number of SN & 200 \\
\quad Capacity & 15 \\
\quad Connection Distance & 20 \\
\quad SN Connect Param. & -85 \\
\quad Beta ($\beta$) & 55 \\
\quad Gamma ($\gamma$) & 2.5 \\
\addlinespace
\midrule
Iterative Parameter & \\
\quad Search Agents & 50 \\
\quad Max Iterations & 100 \\
\bottomrule
\end{tabular}
\caption{Parameter settings for the WSNs problem}
\vspace{0.5cm}

\end{table*}

\begin{table}[htbp]
\centering
\begin{tabular}{@{}lp{5cm}r@{}}
\toprule
Parameter & Parameter Description & Value \\
\midrule
$\varepsilon_{1}$ & Penalty factor for exceeding the number of concepts required for learning  & 1 \\
$\varepsilon_{2}$ & Penalty factor for not covering the number of learning concepts & $10^4$ \\
$\varepsilon_{3}$ & Penalty factor for exceeding the attention span & 1000 \\
$\omega_{1}$ & Coefficient $\varepsilon_{1}$ & 0.25 \\
$\omega_{2}$ & Coefficient $\varepsilon_{2}$ & 1 \\
$\omega_{3}$ & Coefficient $\varepsilon_{3}$ & 0.25 \\
$\psi_{1}$ & High priority material quantity limit & 3 \\
$\psi_{2}$ & Medium priority material quantity limit & 6 \\
$\psi_{3}$ & Challenging material quantity limit & 1 \\
\bottomrule
\end{tabular}
\caption{ACS Constraints}
\vspace{0.5cm}

\centering
\label{tab:wsn_params}
\begin{tabular}{lll}
\toprule
\textbf{Parameter} & \textbf{Value} & \textbf{Description} \\
\midrule
coverage & 10 & Penalty per uncovered sensor node \\
connectivity & 1000 & Penalty for disconnected network \\
power\_std & 100 & Penalty for power standard deviation \\
\bottomrule
\end{tabular}
\caption{WSN Constraints}
\end{table}

\begin{table}[htbp]
\centering
\label{tab:wsn_params}
\begin{tabular}{ll}
\toprule
\textbf{Metaheuristic} & \textbf{Parameter} \\
\midrule
GA &  $pc=0.7,pm=0.1$ \\
PSO &  $w=0.8,c1=2.0,c2=2.0$ \\
SCSO &  $rG \in [2,0]$ \\
SOA & -- \\
WO & $r = 0.4$ \\
\bottomrule
\end{tabular}
\caption{Algorithm Parameters}
\end{table}

\subsection{A.6 Optimal Heuristics for different problems}

MeLA generated the optimal heuristics for the four issues as \textbf{Listing 1} - \textbf{Listing 4}.

\begin{figure}[htbp]  % 使用单栏的figure环境
\centering
\begin{minipage}{\linewidth}  % 限制宽度为单栏
\begin{lstlisting}[language=Python, caption={TSP}, label=code:heuristic1, frame=tb, numbers=left, captionpos=b]
def heuristics_v2(distance_matrix):
    #EVOLVE-START
    eps = 1e-8
    neigh_mean = np.mean(np.sort(distance_matrix, axis=1)[:,1:5], axis=1)
    norm_term = (neigh_mean[:,None] + neigh_mean)/np.maximum(distance_matrix, eps)**2
    return np.reciprocal(distance_matrix + eps) * norm_term
    #EVOLVE-END       
    return 1 / distance_matrix
\end{lstlisting}

\begin{lstlisting}[language=Python, caption={BPP}, label=code:heuristic2, frame=tb, numbers=left, captionpos=b]
def heuristics_v2(node_attr, node_constraint):
    #EVOLVE-START
    n = node_attr.shape[0]
    attr_sum = node_attr[:, None] + node_attr[None, :]
    constraint_diff = np.maximum(1e-6, abs(node_constraint - attr_sum))
    weights = 1 / (1 + constraint_diff * node_attr.mean())
    #EVOLVE-END
    return weights
    
\end{lstlisting}

\begin{lstlisting}[language=Python, caption={ACS}, label=code:heuristic2, frame=tb, numbers=left, captionpos=b]
def heuristics_v2(Positions, Best_pos, Best_score, rg):
    SearchAgents_no = Positions.shape[0]
    dim = Positions.shape[1]

    lb_array = np.zeros((SearchAgents_no, dim))
    ub_array = np.ones((SearchAgents_no, dim))

    rand_adjust = lb_array + (ub_array - lb_array) * np.random.rand(*Positions.shape)
    Positions = np.where((Positions < lb_array) | (Positions > ub_array), rand_adjust, Positions)

    #EVOLVE-START
    beta = 1.5
    sigma = (np.math.gamma(1+beta)*np.sin(np.pi*beta/2)/(np.math.gamma((1+beta)/2)*beta*(2**((beta-1)/2))))**(1/beta)
    levy_step = 0.01 * np.random.randn(SearchAgents_no,1) * sigma / (np.abs(np.random.randn(SearchAgents_no,1))**beta)
    
    learn_prob = 0.5 + 0.4*(Best_score - np.min(np.linalg.norm(Positions-Best_pos,axis=1)))/Best_score
    mask = np.random.rand(SearchAgents_no,dim) < learn_prob.reshape(-1,1)
    Positions = np.where(mask, 
                        Best_pos + levy_step*(Positions - Best_pos.mean(axis=0)), 
                        Positions*(1 + 0.5*(np.random.rand(*Positions.shape)-0.5)))
    #EVOLVE-END       
    return Positions
\end{lstlisting}

\end{minipage}
\end{figure}

\begin{figure}[t]  % 使用单栏的figure环境
\centering
\begin{minipage}{\linewidth}  % 限制宽度为单栏
\begin{lstlisting}[language=Python, caption={WSN}, label=code:heuristic2, frame=tb, numbers=left, captionpos=b]
def heuristics_v2(Positions, Best_pos, Best_score, rg):
    SearchAgents_no = Positions.shape[0]
    dim = Positions.shape[1]

    lb_array = np.zeros((SearchAgents_no, dim))
    ub_array = np.ones((SearchAgents_no, dim))

    rand_adjust = lb_array + (ub_array - lb_array) * np.random.rand(*Positions.shape)
    Positions = np.where((Positions < lb_array) | (Positions > ub_array), rand_adjust, Positions)

    #EVOLVE-START
    # Levy flight component
    beta = 1.5
    sigma = (np.math.gamma(1+beta)*np.sin(np.pi*beta/2)/(np.math.gamma((1+beta)/2)*beta*2**((beta-1)/2)))**(1/beta)
    u = np.random.randn(*Positions.shape) * sigma
    v = np.random.randn(*Positions.shape)
    step = u/abs(v)**(1/beta)
    
    # Adaptive weights
    w = 0.9 - 0.5*(Best_score/1000)  # Scale based on fitness
    
    # Hybrid update
    r = np.random.rand(SearchAgents_no, 1)
    mask = r < 0.5
    Positions = np.where(mask, 
                        Best_pos + w*step*Positions, 
                        w*Positions + (Best_pos - Positions)*np.random.rand(*Positions.shape))
    #EVOLVE-END
    return Positions
\end{lstlisting}
\end{minipage}
\end{figure}
\end{document}